\documentclass[sigconf]{acmart}

\usepackage{booktabs} 
\usepackage{slashbox}
\usepackage{multirow}
\usepackage{caption}
\usepackage{caption}

\copyrightyear{2019}
\acmYear{2019}
\setcopyright{iw3c2w3}
\acmConference[WWW '19 Companion]{Companion Proceedings of the 2019 World Wide Web Conference}{May 13--17, 2019}{San Francisco, CA, USA}
\acmBooktitle{Companion Proceedings of the 2019 World Wide Web Conference (WWW '19 Companion), May 13--17, 2019, San Francisco, CA, USA}
\acmPrice{}
\acmDOI{10.1145/3308560.3316549}
\acmISBN{978-1-4503-6675-5/19/05}

\begin{document}
\title{Multimodal Emotion Classification}

\author{Anurag Illendula}
\affiliation{
  \institution{Department of Mathematics\\ Indian Institute of Technology Kharagpur}
  \city{Kharagpur}
  \state{India}
}
\email{aianurag09@iitkgp.ac.in}

\author{Amit Sheth}
\affiliation{
  \institution{Kno.e.sis Center\\ Wright State University}
  \city{Dayton}
  \state{Ohio}
}
\email{amit@knoesis.org}

\renewcommand{\shortauthors}{Illendula et al.}

\begin{abstract}
Most NLP and Computer Vision tasks are limited to scarcity of labelled data. In social media emotion classification and other related tasks, hashtags have been used as indicators to label data. With the rapid increase in emoji usage of social media, emojis are used as an additional feature for major social NLP tasks. However, this is less explored in case of multimedia posts on social media where posts are composed of both image and text. At the same time, w.e have seen a surge in the interest to incorporate domain knowledge to improve machine understanding of text. In this paper, we investigate whether domain knowledge for emoji can improve the accuracy of emotion classification task. We exploit the importance of different modalities from social media post for emotion classification task using state-of-the-art deep learning architectures. Our experiments demonstrate that the three modalities (text, emoji and images) encode different information to express emotion and therefore can complement each other. Our results also demonstrate that emoji sense depends on the textual context, and emoji combined with text encodes better information than considered separately. The highest accuracy of 71.98\% is achieved with a training data of 550k posts.

\end{abstract}

%
%



\maketitle

\section{Introduction}

Emotion is any conscious experience characterized by intense mental activity and a certain degree of pleasure or displeasure. It primarily reflects all aspects of our daily lives, playing a vital role in our decision-making and relationships. In recent years, there have been a growing interest in the development of technologies to recognize emotional states of individuals. Due to the escalating use of social media, emotion-rich content is being generated at an increasing rate, encouraging research on automatic emoji classification techniques. Social media posts are mainly composed of images and captions. Each of the modalities has very distinct statistical properties and fusing these modalities helps us learn useful representations of the data \cite{barbieri2018multimodal}. Emotion recognition is a process that uses low-level signal cues to predict high-level emotion labels. With the rapid increase in usage of emojis, researchers started using them as labels to train classification models \cite{felbo2017using}. A survey conducted by secondary school students suggested that the use of emoticons can help reinforce the meaning of the message \footnote{\url{https://bit.ly/2NyyiIp}}. Researchers found that emoticons when used in conjunction with a written message, can help to increase the ``intensity'' of its intended meaning \cite{derks2008emoticons}.

\begin{figure}[]
\centering
\includegraphics[width=1.0\linewidth]{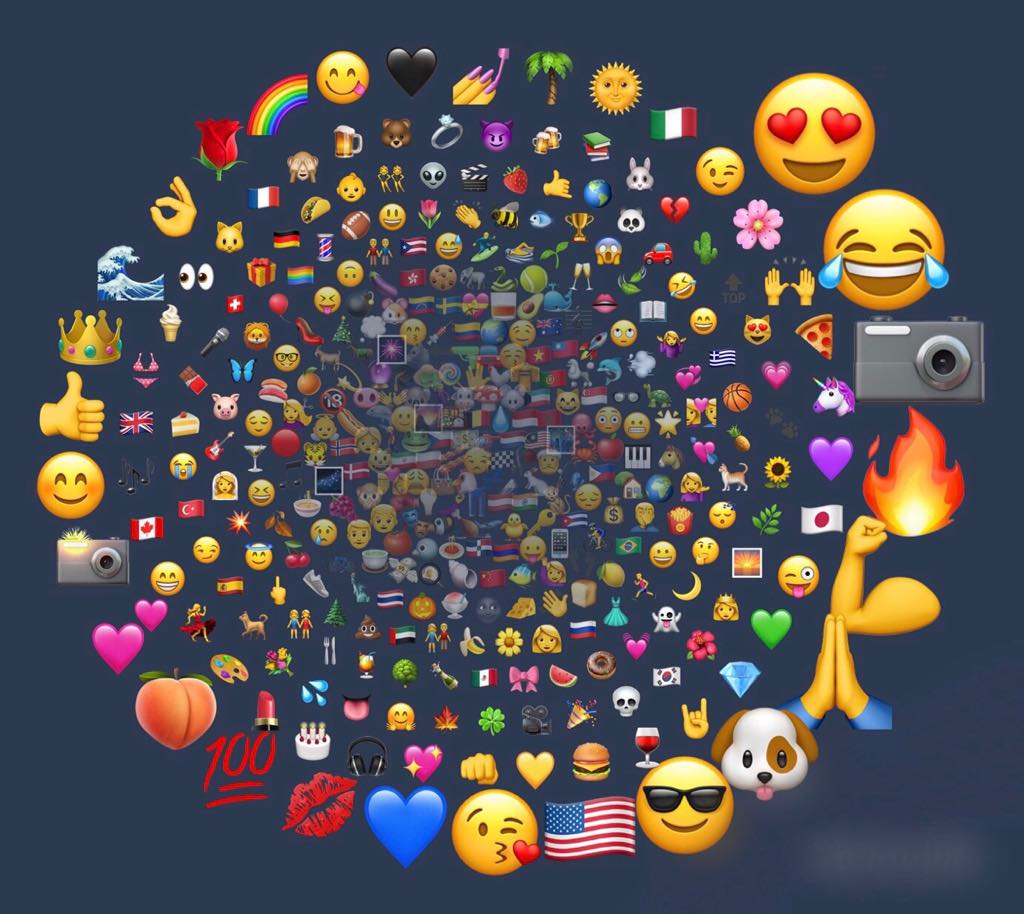}
\caption{Spiral representing usage of different Emojis on Instagram. The image is copied from \url{https://bit.ly/2W3ks5u}, Article by Stefan Pettersson}
\label{fig1}
\vspace{-8mm}
\end{figure}

Emojis are being used for the visual depictions of human emotions \cite{rajhi2017emotional}.  Emotions help us to determine the interactions among human beings. The context of emotions specifically brings out the complex and bizarre social communication. These social communications are identified as the judgment of other person\'s mood based on his emoji usage \cite{rajhi2017emotional}. According to a study made by Rajhi et al. \cite{rajhi2017emotional}, the real-time use of emojis can detect the human emotions in different scenes, lighting conditions as well as angles in real time. Studies have shown that emojis when embedded with text to express emotion make the tone and tenor of the message clearer. This further helps in reducing or eliminating the chances of misunderstanding, often associated with plain text messages\footnote{\url{https://bit.ly/2AV1sNA}}. Recent study proved that co-occurrence helps users to express their sentiment more effectively\footnote{\url{https://bit.ly/2QW91sx}} \cite{illendula2018learning}.

Psychological studies conducted in the early '80s provide us strong evidence that human emotion is closely related to the visual content. Images can both express and affect people\'s emotions. Hence it is intriguing and important to understand how emotions are conveyed and how they are implied by the visual content of images. With this as a reference, many computer scientists have been working to relate and learn different visual features from images to classify emotional intent. Convolutional Neural Networks (CNNs) have served as the baselines for major Image processing tasks. These deep CNNs combine the high and low-level features and classify images in an end-to-end multi layer fashion.

\begin{figure}
\label{fig2}
\centering
\captionsetup{justification=centering}
\includegraphics[width=1.0\linewidth]{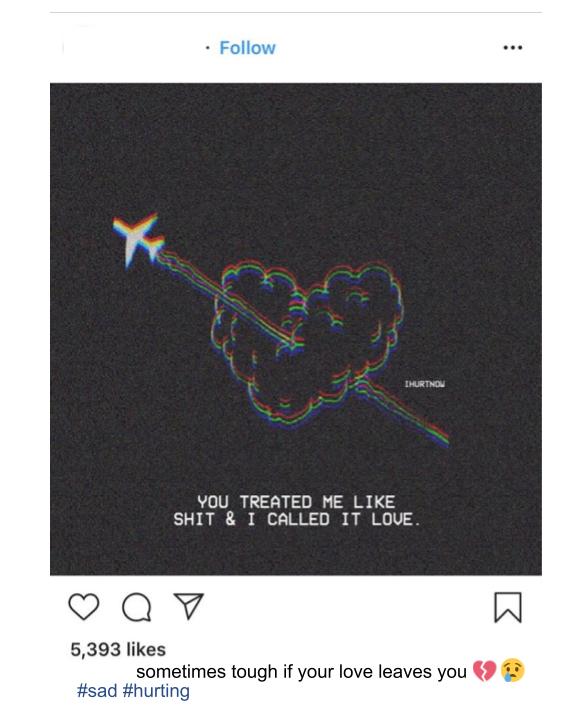}
\caption{Example of an Instagram post which belongs to ``Sad'' emotional category in our dataset}
\vspace{-8mm}
\end{figure}

Earlier most researchers working in the field of social NLP have used either textual features or visual features, but there are hardly any instances where researchers have combined both these features. Recent works by Barbieri et al.'s \cite{barbieri2018multimodal}, Illendula et al.'s \cite{illendula2018emoji} on multimodal emoji prediction and Apostolova et al.’s \cite{apostolova2014combining} work on information extraction fusing visual and textual features have shown that combining both modalities helps in improving the accuracies. While a high percentage of social media posts are composed of both images and caption, the researchers have not looked at the multimodal aspect for emotion classification. Consider the post in \hyperref[fig2]{Firgure 2} where a user is sad and posts an image when a person close to him leaves him. The image represents a disturbed heart and has a textual description ``sometimes tough if your love leaves you \#sad \#hurting'' conveys a sad emotion. Similarly the emoji used \includegraphics[height=1em]{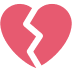}, \includegraphics[height=1em]{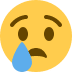} conveys the emotion of being depressed. We hypothesize that all the modalities from a social media post including visual, textual, and emoji features, contribute to predicting the emotion of the user. Consequently, we seek to learn the importance of different modalities towards emotion prediction task. We first discuss relevant research in Section 2. The development of our dataset and present preliminary data analysis experiments which illustrate the importance to study the usage of emojis in different emotional contexts is described in Section 3. We present our model and approach of multimodal emotional classification in Section 4. The results from different experiments and relevant discussion are in Section 5 and Section 6, respectively. We conclude with some interesting findings and our future plan in Section 7.

\section{Related Work}

\begin{table*}[]
\label{table1}
\centering
\captionsetup{justification=centering}
\begin{tabular}{|c|c|c|c|c|c|c|}
\hline
\textbf{Emotion} & \textbf{\begin{tabular}[c]{@{}c@{}}\# of \\ Hash Tags\end{tabular}} & \textbf{\begin{tabular}[c]{@{}c@{}}Examples\\ of Hash Tags\end{tabular}}             & \textbf{\begin{tabular}[c]{@{}c@{}}\# of Instagram\\ Posts with Emoji\end{tabular}} & \textbf{\begin{tabular}[c]{@{}c@{}}\# of Instagram Posts\\      without Emoji\end{tabular}} & \textbf{\begin{tabular}[c]{@{}c@{}}Number of \\ Instagram Posts\end{tabular}} & \textbf{\begin{tabular}[c]{@{}c@{}}Percentage of\\ Posts with Emoji\end{tabular}} \\ \hline
Anger            & 23                                                                  & \begin{tabular}[c]{@{}c@{}}aggravation, irritation,\\  agilation, anger\end{tabular} & 59458                                                                               & 28549                                                                                       & 88007                                                                         & 67.56                                                                             \\ \hline
Fear             & 22                                                                  & \begin{tabular}[c]{@{}c@{}}shock, fear, fright,\\  terror, panic\end{tabular}        & 44511                                                                               & 28078                                                                                       & 63644                                                                         & 61.32                                                                             \\ \hline
Joy              & 36                                                                  & \begin{tabular}[c]{@{}c@{}}excited, happy \\ elated, proud\end{tabular}       & 55566                                                                               & 29947                                                                                       & 85513                                                                         & 64.98                                                                             \\ \hline
Love             & 17                                                                  & \begin{tabular}[c]{@{}c@{}}affection, love, \\ loving, fondness\end{tabular}         & 54156                                                                               & 31596                                                                                       & 89486                                                                         & 56.38                                                                             \\ \hline
Sad              & 36                                                                  & \begin{tabular}[c]{@{}c@{}}sorrow, unhappy, \\ depressing, lonely\end{tabular}       & 54156                                                                               & 31596                                                                                       & 85752                                                                         & 63.16                                                                             \\ \hline
Surprise         & 5                                                                   & \begin{tabular}[c]{@{}c@{}}amazement, surprise,\\  astonishment\end{tabular}         & 45640                                                                               & 17230                                                                                       & 72870                                                                         & 62.25                                                                             \\ \hline
Thankful         & 2                                                                   & thankfulness, thankful                                                               & 42870                                                                               & 31308                                                                                       & 74178                                                                         & 57.80                                                                             \\ \hline
Overall          & 141                                                                 &                                                                                      & 352649                                                                              & 205746                                                                                      & 558395                                                                        & 63.16                                                                             \\ \hline
\end{tabular}
\vspace{2mm}
\caption{Emotion words used for collecting tweets and the statistics of instagram posts in our dataset after filtration}
\vspace{-7mm}
\end{table*}

Most NLP tasks are limited to the scarcity of labeled data. Earlier many researchers have used manual annotation technique to evaluate their models, but this requires much understanding of the emotional content of all the expressions, is time-consuming, requires much effort and may differ according to one's perspective. This creates a misinterpretation of emotion and effects the accuracies of respective tasks. Hence in most social NLP related tasks namely sentiment analysis and other emotion classification tasks hashtags were used as features for automatically labeling data to corresponding categories \cite{wang2012harnessing, felbo2017using}. However, the rapid increase in the usage of emojis on social media helped researchers use emoticons as features for data labeling. Felbo et al. \cite{felbo2017using} has introduced a transfer learning approach for emotion classification, sentiment analysis, sarcasm identification through emoji prediction on the text.

Using emoji knowledge for sentiment analysis, emotion classification and related tasks is not a new idea. Researchers have started using Emojinet \cite{emojineticwsm} to learn embeddings for sentiment analysis task and have achieved better accuracies than the previous state of the art emoji embeddings \cite{wijeratne2017semantics}. There have been many approaches which use emoji as a feature to classify sentiment on social media using emoji as a feature, Illendula et al. \cite{illendula2018learning} has used emoji co-occurrence has a feature to learn sentiment features of emojis. Also, emojis have also been a very important feature to classify emotional content; previous research has always manually specified which emotional category each hashtag or emoji belongs to \cite{novak2015sentiment}. Prior research work has used theories of emotion such as Ekman's six basic emotions, Pluchtik's eight basic emotions, and other psychology works \cite{mohammad2012emotional,suttles2013distant, shaver1987emotion}. Shaver et al.'s \cite{shaver1987emotion} work which is the most used and most cited work helped to automatically label tweet without any human intervention to take care of biases and misinterpretation to corresponding emotional label and develop an emotion classification model for text \cite{wang2012harnessing}.

Content-Based Information Retrieval (CBIR) is the historical line of research in multimedia tasks. This task usually deals with retrieving images in the dataset that are most similar to the query image. Bag-of-Words representation \cite{zhang2010understanding} has seen a sustained line of research in this task which has been effective up to million sized image datasets. This Information Retrieval task has eventually led to research in image classification which is one of the fundamental challenges in Computer Vision. Convolutional Neural Networks (CNNs) \cite{KrizhevskySH17} have shown promising results in image classification. Research on shortcut connections has been an emerging topic since the development of multilayer perceptrons which has shown promising results for Image processing tasks. Generally, these multiple layers have been connected using shortcut connections using gated functions \cite{hochreiter1997long}. In image classification, depth of the network, i.e., number of layers within the network is of crucial importance as noted by Simonyan et al. \cite{Simonyan14c}. Increasing the depth can have an adverse effect on the image classification task. Most notably, vanishing/exploding gradients problem~\cite{pmlr-v9-glorot10a} and the degradation problem~\cite{He2015} are of significant importance. These problems are overcome by the introduction of shortcut connections and residual representations introduced in Residual Networks \cite{He2016} which won the 1st place in ImageNet 2015 Image classification competition \footnote{\url{https://bit.ly/2y4J8Cz}}. Residual Networks and its extensions which consist of many residual units have shown to achieve state-of-the-art accuracy for image classification tasks on datasets such as ImageNet \cite{ILSVRCanalysis_ICCV2013}.

Knowledge graphs have been proved to be on the important addition to data for training machine learning and deep learning model architectures for various Natural language and Image Processing tasks. Recent research in NLP Emojinet has improved the accuracies of emoji similarity \cite{wijeratne2017semantics}, learning emoji embeddings, emoji sense disambiguation \cite{emojineticwsm}. Emojinet has also improved the accuracies of emoji prediction in case of images \cite{illendula2018emoji}. In the case of image processing tasks, knowledge graphs have also improved the task of video captioning \cite{venugopalan2016improving} and other related tasks \cite{marino2016more}.

In this paper, we present an emotion classification approach using techniques in deep learning leveraging the three modalities from social media posts namely textual, visual and emoji features. We use state of the art approaches for learning features from three modalities namely ResNet model architecture, attention mechanism, Bag of words model for visual, textual and image features respectively. We used two different word embeddings namely FastText (word embedding model which is capable of capturing sub word information) \cite{joulin2016fasttext} model trained on the set of the processed captions (we explain the pre-processing steps involved in Section 3) and pretrained fast text model trained on Wiki corpus \cite{mikolov2018advances}. We use the bag of words model developed by Wijeratne et al. \cite{wijeratne2017semantics} to learn the emoji embeddings using the emoji knowledge concepts extracted from Emojinet. We report the results observed considering different emoji knowledge concepts from  EmojiNet namely emoji names, emoji senses, emoji sense definitions, and three different word embedding models. We also discuss the importance of different modalities towards emotion classification and report our results and observations in further sections.

\section{Dataset Development}

\subsection{Instagram Dataset}

Instagram is one of the major platforms where people share multimedia posts composed of images and text in the form of description. Hence Instagram provides us an emotion-rich content to analyze different user emotions using machine learning and deep learning techniques. We have used the same approach as Wang et al. \cite{wang2012harnessing} to collect labeled data. Following Shaver et al. \cite{shaver1987emotion} they collected set of possible hashtags for six different emotions (for example words amazement, surprise, astonishment convey the emotion ``Surprise''). This approach to collect emotion-labeled posts has been proved to be effective to label text. Shaver et al. have constructed an emotion-related tree where the first layer represents the parent category, i.e., the six different emotion, the second layer consists of 25 emotion-related keywords categorized into six different parent emotions. We have added a seventh emotional category different from the above mentioned emotional categories named ``Thankful''. We then checked posts having hashtags \#gratified,  \#humbled, \#blessed, \#thankful on instagram, because in general these key words mean thankful according to english literature. But there are hardly any posts having a hashtag \#gratified (around 4k posts in total and no posts in our analysis period). We then checked posts having hashtags \#humbled and \#blessed but several posts convey a Joy emotion rather than Thankful emotion. Prior research of study of emotion related key words convey that \# blessed could be considered as happy emotion but not necessarily Thankful or Joy or Love emotion is particular \cite{liew2016exploring}. Hence we considered only the key words reported by Wang et al. for the seventh emotional category namely \#thankful and \#thankfulness for Thankful emotion.  We then followed the pre-processing steps for emotion-related keywords by adding lexical variants of a word and removing ambiguous words. \hyperref[table1]{Table 1}  gives a complete idea of the set of hashtags along with the number of posts collected from Instagram. We then used Instalooter\footnote{\url{https://bit.ly/2RGBYxF}} API which is an open source tool to collect public posts from Instagram. Instalooter API gives us access to HashTaglooter method similar to Twitter's hashtag search which helps us to collect multimedia posts having a particular hashtag. For example, we collected the posts having ``\#amazement'' and labeled the post to Surprise emotional category. Similarly, we extracted the emotion-labeled multimedia posts for other emotion labels.

\begin{figure}[]
\centering
\captionsetup{justification=centering}
\label{fig3}
\centering
\includegraphics[width=1.0\linewidth]{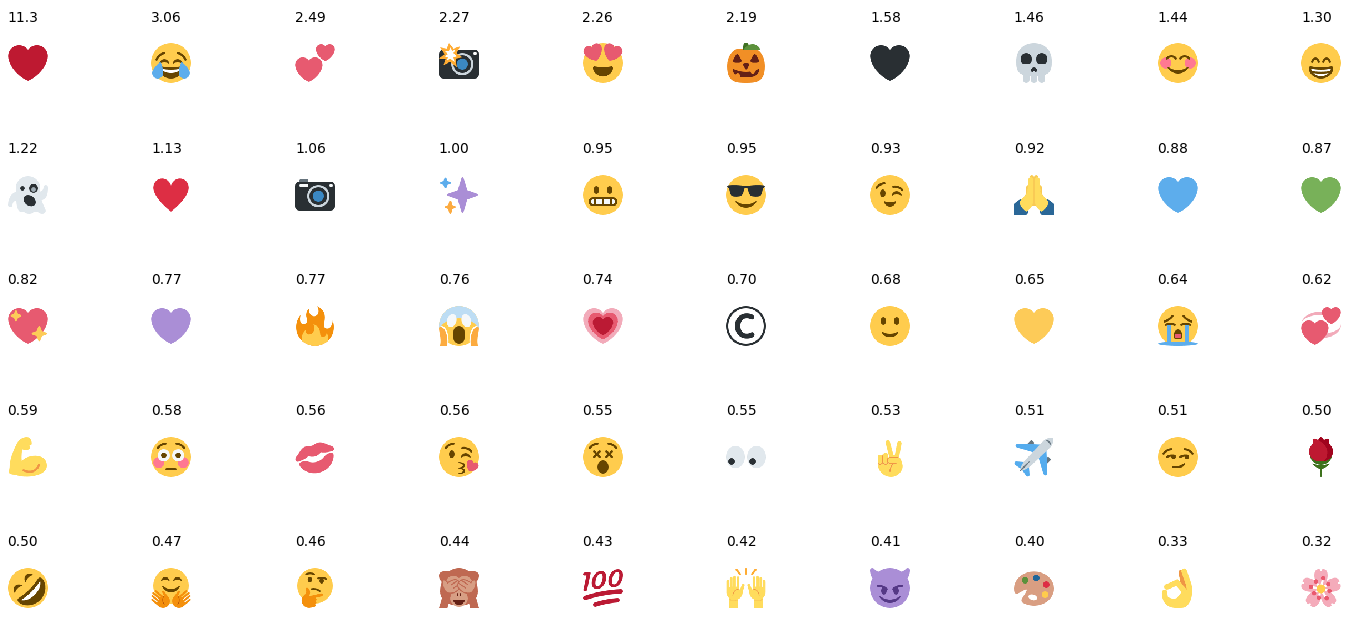}
\caption{Percentage of number of posts having different types of emojis embedded in their post description.}
\end{figure}

\subsection{Pre Processing}

In this section, we explain the different pre-processing steps to ensure the quality of our Instagram dataset. In total, we have collected about 1.1 million posts of different languages from Instagram using instalooter API posted between 1st December 2018 to 20th December 2018, a duration of 20 days. The preprocessing of this corpus consisted of a set of filtering, followed by annotation, as discussed next.

\noindent\textbf{Filtering:}
\begin{itemize}
\item Our current approach is limited to English language only, so we filtered the dataset for the posts that used English language only. 
\item We also ensured that the multimedia posts contain only images but not videos. Instagram also allows users to post multiple images in the same post, so we have filtered them and also considered posts containing one image but not multiple images. 
\item The next step of filtration involved removal of posts which have multiple hashtags belonging to different emotional content. For instance, user's have used surprise and joy in the same post, and we couldn't categorize this post to a single emotional category. 
\item We also made sure that we remove all the emotion-related hashtags, for instance, users may have used \#amazement and \#surprise in the same post. Hence we removed all the emotion-related hashtags which are present in our lexicon. 
\item We then removed posts having URL's embedded in the textual description section.  We hypothesize that the content obtained from the URLs are likely to be important for understanding the emotional content of the post. 
\end{itemize}

\noindent\textbf{Annotation and Text Processing:} 
\begin{itemize}
\item Each post was annotated first by removing the emotion-related keywords from the textual description with the corresponding emotion label. 
\item Suppose, if the description of the posts has \#bliss, then we removed the emotion-related hashtag and then labeled the post to ``Joy'' emotional category, if a hashtag is not an emotion-related keyword we converted the hashtag to corresponding word (``\#car'' to ``car''). 
\item We also followed some text pre-processing steps like removing characters repeated more than two times in a word (e.g., looooool to lol). 
\item We replaced user mentions to ``USER'' to anonymize users. 
\end{itemize}

\subsection{Explorational Data Analysis}

Leveraging emoji as an additional feature has improved accuracies of various socialNLP tasks. Examples include improved user classification based on marijuana usage \cite{kursuncu2018s}, better twitter street gang member identification \cite{balasuriya2016finding}, and higher accuracy in sarcasm and emotion identification on Twitter \cite{felbo2017using}. A recent study by Scott Ayres on emoji usage on the Instagram platform reports that 31\% of all image posts and 29\% of all video posts contained emojis \footnote{\url{https://bit.ly/2E7rO1L}}. Hence, we have analyzed the usage of emojis in each emotional category. \hyperref[table1]{Table 1} represents the number of posts having emojis embedded in the textual description out of the complete set of posts under each emotional category. We observed that about 63.16\% of posts contain emojis. In many cases, users use them to express their emotion which cannot be conveyed in the form of text. Hence we explore the use of domain knowledge of emojis from EmojiNet\footnote{\url{https://bit.ly/2HiSRIq}} in our multimodal emotion classification task and reported our results in further sections. 

\hyperref[fig3]{Figure 3} shows the percentage of posts having different emojis. We have found that users used more than 1800 different types of emojis in their posts, the emoji \includegraphics[height=1em]{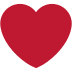} which is the most frequently used emoji, is used in only 11.3\% posts, which is about ${1/9^{th}}$ of the total number of posts. We have then looked into the use of emojis to express a particular emotion. \hyperref[table2]{Table 2} reports the five most frequently used emojis to express a particular emotion. Consequently we looked at the sense forms of these emojis from EmojinNet, and the results looked convincing. For instance, the most frequently used emojis under the emotional category ``Sad'', \includegraphics[height=1em]{1679.png},  \includegraphics[height=1em]{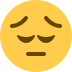}, \includegraphics[height=1em]{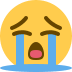}, \includegraphics[height=1em]{956.png}, \includegraphics[height=1em]{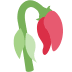} have sense forms which are closely related to sad or depressed. Similarly, we have studied emoji usage in other emotional categories and observed that the emojis corresponding to each emotional category have sense forms related to the emotional category. This supports our hypothesis that emojis play a vital role in relating to users' emotion.

\begin{table}[]
\label{table2}
\centering
\captionsetup{justification=centering}
\begin{tabular}{|c|c|c|l|l|l|}
\hline
\backslashbox{Emotion}{Emojis} & 1 & 2 & 3 & 4 & 5 \\ \hline
Anger    & \includegraphics[height=1em]{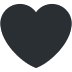}   & \includegraphics[height=1em]{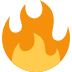}  & \includegraphics[height=1em]{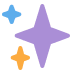}  & \includegraphics[height=1em]{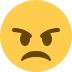}  & \includegraphics[height=1em]{1679.png}  \\ \hline
Fear     & \includegraphics[height=1em]{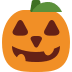}   & \includegraphics[height=1em]{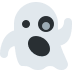}  & \includegraphics[height=1em]{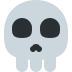}  & \includegraphics[height=1em]{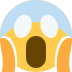}  & \includegraphics[height=1em]{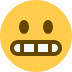}  \\ \hline
Joy      & \includegraphics[height=1em]{1268.png}   & \includegraphics[height=1em]{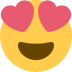}  & \includegraphics[height=1em]{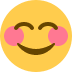}  & \includegraphics[height=1em]{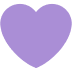}  & \includegraphics[height=1em]{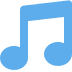}  \\ \hline
Love     & \includegraphics[height=1em]{1268.png}   & \includegraphics[height=1em]{979.png}  & \includegraphics[height=1em]{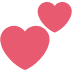}  & \includegraphics[height=1em]{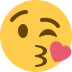}  & \includegraphics[height=1em]{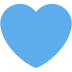}  \\ \hline
Sad      & \includegraphics[height=1em]{1679.png}   & \includegraphics[height=1em]{971.png}  & \includegraphics[height=1em]{943.png}  & \includegraphics[height=1em]{956.png}  & \includegraphics[height=1em]{73.png}  \\ \hline
Surprise & \includegraphics[height=1em]{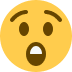}   & \includegraphics[height=1em]{979.png}   & \includegraphics[height=1em]{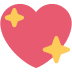}  & \includegraphics[height=1em]{982.png}  & \includegraphics[height=1em]{1343.png}  \\ \hline
Thankful & \includegraphics[height=1em]{1268.png}   & \includegraphics[height=1em]{1678.png}  & \includegraphics[height=1em]{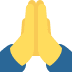}  & \includegraphics[height=1em]{982.png}  & \includegraphics[height=1em]{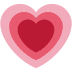}  \\ \hline
\end{tabular}
\vspace{4mm}
\caption{The five most frequently used emojis to express a particular emotion}
\vspace{-10mm}
\end{table}

\section{Pre Training}

We collected set of all user captions from each post and followed pre-processing steps and trained a fastText word embedding model. We chose fasttext over other conventional word embedding models due to its capability of capturing sub-word information which plays a significant role in social NLP tasks. Also fasttext word embedding model has proved to give greater accuracies compared to other word embedding models over various NLP tasks such as sentiment analysis, emotion detection, sarcasm identification \cite{joulin2016fasttext}, and emoji prediction \cite{illendula2018emoji}.

As we have observed that the emotional intent of the emoji used and the emotion expressed by the user in the post are parallel, we tried to use EmojiNet which lists 21,000 emoji sense definitions and 4,618 emoji sense forms to 2,389 emojis. Emoji sense forms list the possible sense forms related to each emoji and emoji sense definitions explaining the context of use of emojis \cite{emojineticwsm}. We chose to use the Bag of words model developed by Wijeratne et al. \cite{wijeratne2017semantics} to learn the emoji representations in the same vector space as words.

\section{Models}

In this section, we present and motivate the models that we used to learn features from an Instagram post composed of a picture and the associated comment for emotion classification task.

\subsection{ResNets}

ResNet architecture \cite{He2016}, composed of convolution layers is currently the state of the art for image classification tasks achieving high accuracies\cite{ILSVRCanalysis_ICCV2013}. Before the introduction of ResNet's, deep CNNs were used for most image processing tasks. ResNet is a feed forward CNN that utilizes two or more convolution layers \cite{sermanet2011traffic}. It has been observed that the accuracy of image classification majorly depends on the depth of the network, i.e., the number of layers which can be related to the number of layers within the network \cite{pmlr-v9-glorot10a}. Hence we used different types of ResNet model architectures, namely ResNet-101 and ResNet-152 to check the classification accuracy in different cases. ResNet model architecture was first tested on the ImageNet dataset where the number of classes for classification was 1000. Since our task of emotion classification requires us to learn features from both text and image, we use the ResNet model to learn a feature vector from an image. We use the implementation of ResNet by Kotikalapudi which is open-sourced and available on github\footnote{\url{https://bit.ly/2W69fRC}}. \hyperref[table3]{Table 3} reports the precision, recall, macro F1 score observed using different ResNet model architectures.

\begin{figure}[]
\label{fig4}
\captionsetup{justification=centering}
\centering
\includegraphics[width=1.0\linewidth]{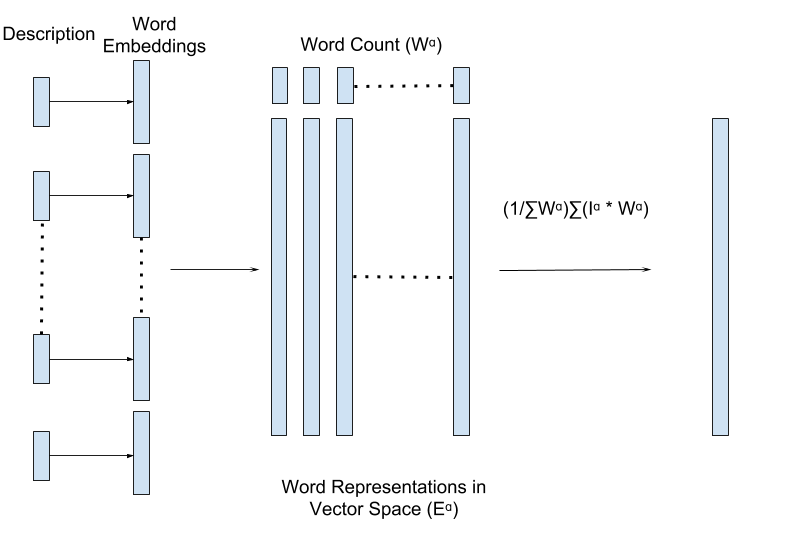}
\caption{Bag Of Words model to learn emoji representation in the similar vector spaces as words.}
\end{figure}

\subsection{Bag Of Words Model}

As discussed earlier there are different word embedding models which learn rich representations of words in vocabulary using neural networks. The neural network takes a large n-dimensional vector for each word (where n is the number of distinct words in the vocabulary), then learns a transformation of the vector in a low dimensional vector space. There have been many word embedding models to learn word representations including GloVe, skip-gram and CBOW. Similarly, there are different approaches to learn emoji embeddings that use co-occurrence feature \cite{illendula2018learning}, skip-gram word embedding architecture \cite{barbieri2016does}, and semantic knowledge of emojis from EmojiNet \cite{wijeratne2017semantics}. Since EmojiNet gives us access to the emoji sense forms, we make use of the embedding model developed by Wijeratne et al. \cite{wijeratne2017semantics} to learn emoji embeddings.  We also check the accuracy of our approach using other emoji embeddings. Wijeratne et al. \cite{wijeratne2017semantics} replaced the word vectors of all words in the emoji definition and formed a 300-dimensional vector performing vector average. Also, the vector mean (or average) adjusts for word embedding bias that could take place due to certain emoji definitions having considerably more words than others has been noted by Kenter et al. \cite{kenter2016siamese}.

We make use of different knowledge concepts, namely sense forms, sense definitions, emoji names from EmojiNet and use the bag-of-words model (\hyperref[fig4]{Figure 4}), fastText trained word embeddings to learn the emoji representations. Since we use the word embeddings to learn emoji representation, we could say that both emojis and words are embedded on a similar vector space.  We define two types of knowledge embeddings termed as Emoji\_Embeddings \_Definitions, Emoji\_Embeddings\ SenseLabels learned using emoji sense definitions and emoji sense forms extracted from EmojiNet respectively. Then we evaluate our model using these embeddings as external knowledge concepts.

Emoji\_Embeddings \_Definitions: Emoji Definitions are the textual descriptions that relate to the context of use of particular emoji. The emoji embedding from the set of descriptions is calculated using the bag-of-words model shown in \hyperref[fig4]{Figure 4}. For example, consider the emoji \includegraphics[height=1em]{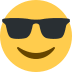},  EmojiNet lists ``One of the temperate seasons, Summer is the warmest of the four temperate seasons, falling between spring and autumn.'', ``The period or season of summer.'' and so forth as emoji definitions for the emoji (\includegraphics[height=1em]{977.png}).

Emoji\_Embeddings \_SenseLabels: Emoji Sense labels are the list of different senses what emoji mean in different contexts. The emoji embedding from the set of Sense labels is calculated following the bag-of-words model shown in \hyperref[fig4]{Figure 4}. For example ``amusing'', ``swagger'' and so forth are the sense labels listed by EmojiNet for the emoji (\includegraphics[height=1em]{977.png}).

Let $C_i$ represent the word count of word $W_i$, $E_i$ represent the word embedding of word $W_i$; then the emoji embedding can be calculated as :

\[Emoji\_Embedding = \frac{\sum E_i*C_i}{\sum C_i}\]

\subsection{Attention Model}

Recurrent neural networks (RNNs) are a class of neural networks that take sequential data or time series data as input and compute hidden state vector at each time step, and these networks make use of the entire history of inputs to compute hidden state vector. A LSTM network is a special class of RNNs which has a memory cell and three gating units. These three gating units -- input gate, forget gate, and output gate, allow the model to control what information to add, what information to use from previous history, and what information to use to output from current memory cell, respectively. Each gate, implemented as a logistic function $\sigma$, takes input vector and outputs a value between 0 and 1.

Bi-directional LSTM's are special kind of LSTM's which combine two LSTM running in forward and backward directions. The hidden state vector $\vec h_t$ is computed by the forward LSTM, $\overleftarrow h_t$ represent the hidden state vector computed by the backward LSTM. Then we compute the hidden state vector $h_t$ concatenating both the hidden state vector, $h_t = [\vec h_t: \overleftarrow h_t]$. These networks have been shown to be efficient for a wide range of NLP tasks and have improved the accuracy over the LSTM networks because the current hidden state vector is computed using the past and future information. The importance of a word is highly context-dependent, i.e., the same word can have a different degree of importance in different context. To incorporate this perspective, we add an attention layer \cite{yang2016hierarchical} on top of this encoded bidirectional LSTM which helps the model decide the importance of each word for the emotion classification task. For instance, words like ``lovely'' or ``extraordinary'' are likely to add more weight to the emotion carried in the text. The attention layer helps the model learn the importance of each word using attention scores.

\begin{figure}[]
\label{fig5}
\captionsetup{justification=centering}
\centering
\includegraphics[width=1.0\linewidth]{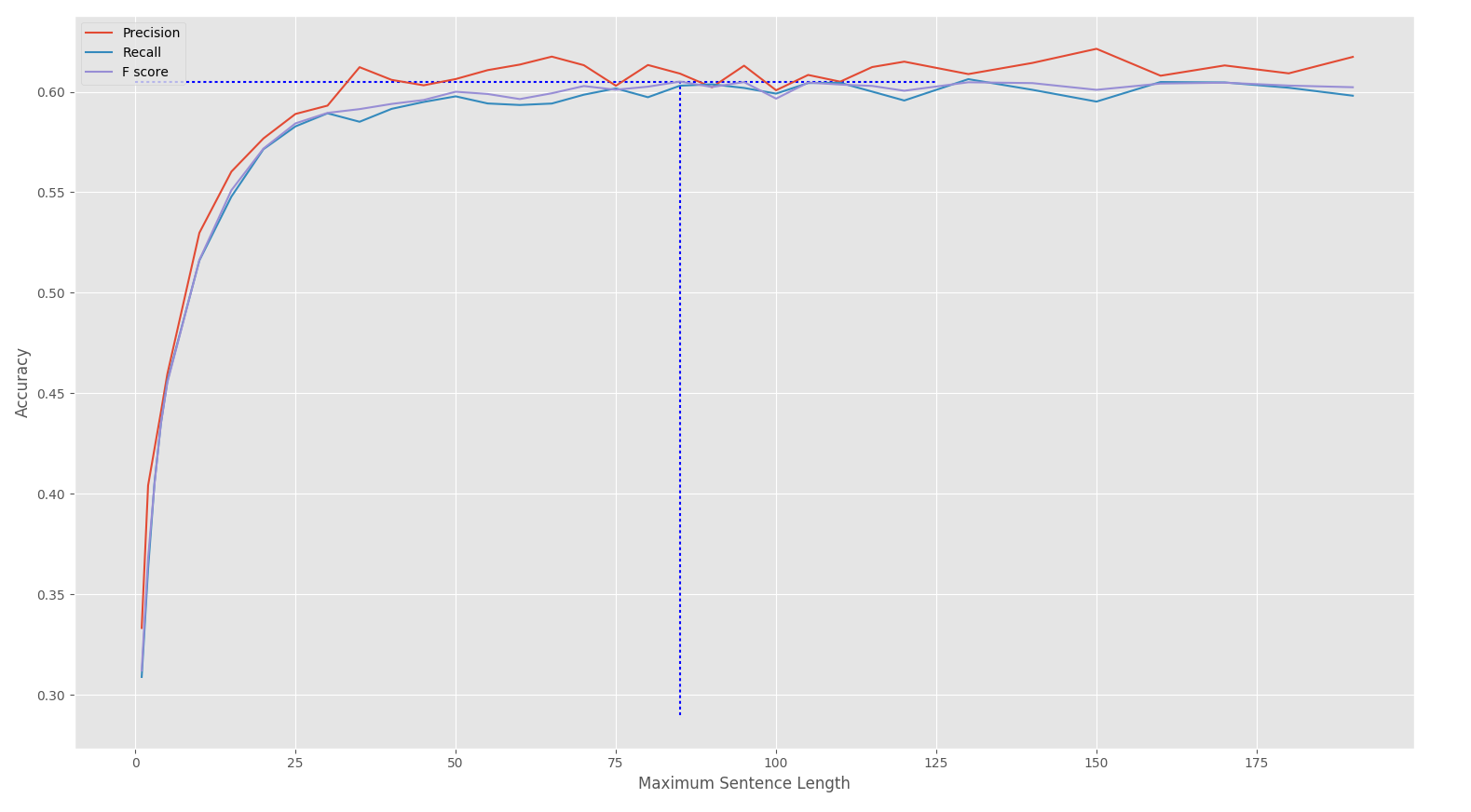}
\caption{Variation of Accuracy measures vs the input sentence length}
\end{figure}

\section{Experiments}

In this section, we report the accuracies of different experiments which we performed. We first check the classification accuracy when each modality is used independently, then we combine these modalities and check the classification accuracy of multimodal features towards the emotion classification task. All models are trained using the Keras library and run on Theano background on a cuda GPU, using Adam's gradient descent optimizer. For all experiments, models are trained using 60\% of the dataset, validated on 20\% of the dataset and tested on the remaining 20\% of the dataset.

\subsection{Visual Features}

We have used the 152-layered residual network which is the current state of the art model architecture for image classification. Also the state of the art model architectures for emoji prediction task for images have used ResNet architecture \cite{barbieri2018multimodal,illendula2018emoji}. Here we trained a 152-layered residual network with learning rate as 0.0001 and stopped the training when there is no further increase in accuracy on the validation set. However, we also checked the classification accuracy using other ResNet model architectures.

\begin{table}[]
\label{table3}
\centering
\captionsetup{justification=centering}
\begin{tabular}{|c|c|c|c|}
\hline
\textbf{Resnet Architecture} & \textbf{Accuracy} & \textbf{Precision} & \textbf{F score} \\ \hline
Resnet-101           & 25.2     & 25.8      & 25.4    \\ \hline
Resnet-152          & \textbf{29.6}     & \textbf{29.8}      & \textbf{29.7}    \\ \hline
\end{tabular}
\vspace{5mm}
\caption{Accuracies of  emotion classification task using visual features from images}
\vspace{-10mm}
\end{table}

\subsection{Emoji features}

As seen in \hyperref[table2]{Table 2}, emojis play an integral part in predicting user's emotions on social media. As discussed earlier, we use the Bag of Words model developed by Wijeratne et al. \cite{wijeratne2017semantics} to learn emoji embeddings using emoji knowledge concepts extracted from EmojiNet. We learn four different emoji embeddings using emoji names, emoji sense forms, emoji sense definitions, and processed emoji sense definitions. Processed emoji sense definitions are the sense definitions formed after following pre-processing steps, namely removal of stop words, lemmatizing words on the emoji sense definitions extracted from EmojiNet. We consider the vector average of the emoji embeddings of emojis emb2dded in each post as the emoji feature vector for each post. For example if a post is embedded with the emoji set (\includegraphics[height=1em]{1679.png}, \includegraphics[height=1em]{1679.png}, \includegraphics[height=1em]{971.png}), then we take $Feature\_Vector$ = $1/3$(vec(\includegraphics[height=1em]{1679.png}) + vec(\includegraphics[height=1em]{1679.png}) + vec(\includegraphics[height=1em]{971.png})). Next we train a neural network using these feature vectors as input and encoded emotion vector as labels.

\begin{table}[H]
\label{table4}
\centering
\captionsetup{justification=centering}
\begin{tabular}{|c|c|c|c|}
\hline
\textbf{\begin{tabular}[c]{@{}c@{}}Knowledge Concept\\ Embeddings\end{tabular}} & \textbf{Precision} & \textbf{Recall} & \textbf{\begin{tabular}[c]{@{}c@{}}Macro F1\\ Score\end{tabular}} \\ \hline
\begin{tabular}[c]{@{}c@{}}Fasttext trained\\ Emoji Embeddings\end{tabular}                                                                     & 20.13             & 19.87           & 19.43                                                            \\ \hline
Emoji Names                                                                     & 27.55              & 26.67           & 25.83                                                             \\ \hline
Emoji Sense Forms     & \textbf{34.81}              & \textbf{30.82}           & \textbf{32.83}                                                             \\ \hline
Emoji Sense Definitions                                                         & 29.16              & 27.06           & 27.38                                                             \\ \hline
\begin{tabular}[c]{@{}c@{}}Processed Emoji\\ Sense Definitions\end{tabular}                                                               & 30.39              & 28.56           & 29.66                                                             \\ \hline
\end{tabular}
\caption{Accuracies of emotion classification task using emoji features from caption and fasttext trained word embeddings on the set of descriptions}
\vspace{-5mm}
\end{table}

\vspace{-8mm}

\subsection{Textual Features}

It has been proved that textual information is more appropriate than visual features and increases the accuracy of emoji prediction in case of multimodal information \cite{barbieri2018multimodal}. Hence we tried to check the importance of textual features for emotion classification task. We use the attention mechanism to learn the importance of words towards emotion classification task. Instagram allows users to post a caption of maximum 2200 characters which translates to approximately 300 words \footnote{\url{https://bit.ly/2RAg2zu}}. It would be challenging to train a Bi-LSTM with such long input sentence length, we experimented on different input sentence lengths to check the variation of accuracy of Bi-LSTM on sentence length. \hyperref[fig5]{Figure 5} gives the clear picture of variation of accuracy measures namely precision, recall and F score with input sentence length to the Bi-LSTM using word embeddings trained using fasttext on the set of captions. We have observed that maximum accuracy for classification is achieved when the input sentence length is nearly 80 words per caption and thereafter there is no significant rise in the accuracy of the model. We also reported our accuracies of the emotion classification task using the bag of words model (which is also a extensively used approach for long input sentences) for text using different word embeddings, and we have observed that attention mechanism has achieved better accuracies than the bag of words model. This can be due to the reason that bag of words model gives equal importance to all the words which is not the case with attention mechanism. \hyperref[table5]{Table 5} and \hyperref[table6]{Table 6} report the accuracies for emotion classification task using different word embeddings and using different model architectures attention mechanism and bag of words model respectively.

\subsection{Combining both Textual and Emoji features}

\begin{figure}[]
\label{fig6}
\centering
\captionsetup{justification=centering}
\includegraphics[width=1.0\linewidth]{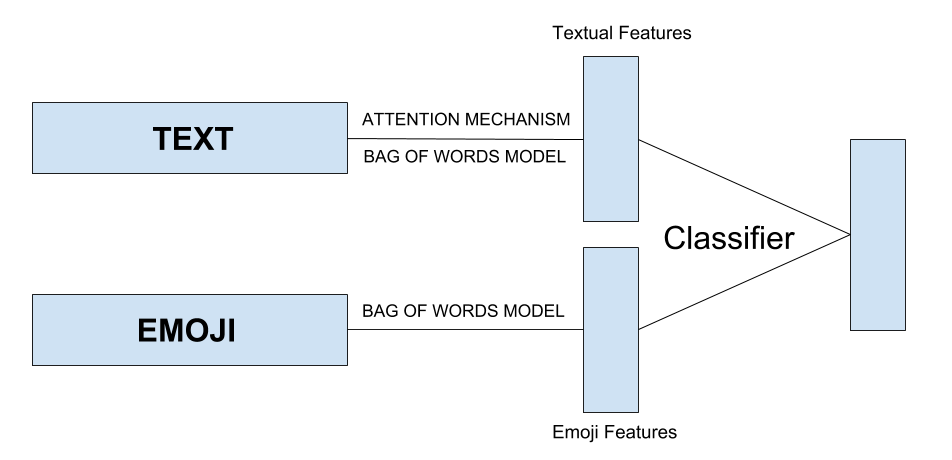}
\caption{Multimodal emotion classification approach using caption of a post.}
\end{figure}

Here we present two different model architectures: one which considers complete caption (emojis embedded in between text) as sequential input and fed to a single input layer, while the other considers text and emoji as different input features and are fed to two different input layers as illustrated in \hyperref[fig6]{Figure 6}.

\begin{table}[]
\label{table5}
\centering
\captionsetup{justification=centering}
\begin{tabular}{|c|c|c|c|c|}
\hline
\textbf{\begin{tabular}[c]{@{}c@{}}Caption\\ Length\end{tabular}} & \textbf{Word Embedding Model} & \textbf{Precision} & \textbf{Recall} & \textbf{\begin{tabular}[c]{@{}c@{}}Macro F1\\ Score\end{tabular}} \\ \hline
20 & \begin{tabular}[c]{@{}c@{}}Fasttext trained \\ on Post Descriptions\end{tabular} & 57.67              & 57.14           & 57.17                   \\ \hline
20 & \begin{tabular}[c]{@{}c@{}}Pretrained Fasttext\\ on Wiki Corpus\end{tabular}     & 41.84               & 38.92           & 40.24                   \\ \hline
40 & \begin{tabular}[c]{@{}c@{}}Fasttext trained on\\ Posts Descriptions\end{tabular} & 60.59               & 59.15           & 59.39                                 \\ \hline
40 & \begin{tabular}[c]{@{}c@{}}Pretrained Fasttext \\ on Wiki Corpus\end{tabular}    & 46.83                & 42.39           & 43.20                                   \\ \hline
80 & \begin{tabular}[c]{@{}c@{}}Fasttext trained on\\ Posts Descriptions\end{tabular} & 61.33               & 59.72             & 60.25                                 \\ \hline
80 & \begin{tabular}[c]{@{}c@{}}Pretrained Fasttext \\ on Wiki Corpus\end{tabular}    & 45.95                 & 44.88           & 43.05                                   \\ \hline
\end{tabular}
\vspace{5mm}
\caption{Accuracies of emotion classification task using textual features from text and attention mechanism at different caption length}
\vspace{-5mm}
\end{table}

\begin{table}[]
\label{table6}
\centering
\captionsetup{justification=centering}
\begin{tabular}{|c|c|c|c|}
\hline
\multicolumn{1}{|l|}{\textbf{Word Embedding Model}}                              & \multicolumn{1}{l|}{\textbf{Precision}} & \multicolumn{1}{l|}{\textbf{Recall}} & \multicolumn{1}{l|}{\textbf{F1 Score}} \\ \hline
\begin{tabular}[c]{@{}c@{}}Fasttext trained on\\ Posts Descriptions\end{tabular} & 56.57                            & 55.82                                & 54.50                                 \\ \hline
\begin{tabular}[c]{@{}c@{}}Pretrained Fasttext \\ on Wiki Corpus\end{tabular}    & 44.26                                    & 43.57                               & 43.47                                   \\ \hline
\end{tabular}
\vspace{5mm}
\caption{Accuracies of emotion classification task using textual features from text and Bag Of Words model}
\vspace{-5mm}
\end{table}

\subsubsection{Bag Of Words approach for whole caption}

Here we consider complete caption as the input to the classification model, and learn the feature vector by calculating vector average of the embeddings of entities, i.e., word embeddings of words and emoji embeddings of emojis. Since the emoji embeddings are learned using the word embeddings, this supports the addition of word embeddings and emoji embeddings to calculate vector average. \hyperref[table8]{Table 8} reports the accuracies of emotion classification task using bag of words model with different emoji knowledge concepts and different word embedding models.

\subsubsection{Attention mechanism for whole caption}

According to the study by Kevin Cohn which says emojis can be used as language on social media\footnote{\url{https://bit.ly/2UbmTRG}} it is noted that the sense of emoji depends on the context of use \cite{emojineticwsm}. Hence Bi-LSTM which have been the SOTA for sequential data would capture the sense of the emoji in the textual context with the help of recurrent units. Hence we train an attention model to check the accuracy where complete caption (emoji+text) is fed to the input layer of a bi-lstm. \hyperref[table7]{Table 7} reports the accuracies of emotion classification task using attention mechanism on whole caption.

\subsubsection{Considering emoji and text as different features}

Recent research by Wijeratne et al. has proved that sense of emoji depends on the textual context \cite{emojineticwsm}. To prove this assertion in the case of emotion classification we use caption without emojis (only text) and emojis embedded in the caption as different features to train classification model. Here we feed the textual input to the attention layer, feature vector learned using emoji embedded in caption to different input layers. We then merge the outputs of attention layer and emoji input layer and add a softmax layer on top of this for classification. The classification accuracies for this model architecture are reported in \hyperref[table9]{Table 9}. \hyperref[fig6]{Figure 6} illustrates the model architecture in this case, where we learn independent features and then merge the features.

\begin{table*}[]
\label{table7}
\centering
\captionsetup{justification=centering}
\begin{tabular}{|c|c|c|c|c|}
\hline
\textbf{\begin{tabular}[c]{@{}c@{}}Word Embedding \\ Model\end{tabular}}                       & \textbf{\begin{tabular}[c]{@{}c@{}}Knowledge\\ Concepts\end{tabular}} & \textbf{Precision} & \textbf{Recall} & \textbf{\begin{tabular}[c]{@{}c@{}}Macro\\ F1 Score\end{tabular}} \\ \hline
\multirow{4}{*}{\begin{tabular}[c]{@{}c@{}}Fasttext on \\ Post Descriptions\end{tabular}}      & Emoji Names                                                           & 66.20              & 64.82           & 65.23                                                             \\ \cline{2-5} 
                                                                                               & Emoji Senses                                                          & 65.40              & 64.96           & 64.95                                                             \\ \cline{2-5} 
                                                                                               & Emoji Sense Definitions                                               & 66.04              & 64.99           & 65.27                                                             \\ \cline{2-5} 
                                                                                               & Processed Emoji Sense Definitions                                     & 65.94              & 64.58           & 65.04                                                             \\ \hline
\multirow{4}{*}{\begin{tabular}[c]{@{}c@{}}Pretrained Fasttext \\ on Wiki Corpus\end{tabular}} & Emoji Names                                                           & 53.99              & 52.14           & 52.42                                                             \\ \cline{2-5} 
                                                                                               & Emoji Senses                                                          & 54.95              & 51.21           & 51.99                                                             \\ \cline{2-5} 
                                                                                               & Emoji Sense Definitions                                               & 52.34              & 51.85           & 51.83                                                             \\ \cline{2-5} 
                                                                                               & Processed Emoji Sense Definitions                                     & 53.72              & 51.83           & 52.25                                                             \\ \hline
\end{tabular}
\vspace{4mm}
\caption{Accuracies of emotion classification task using attention mechanism combining text and emoji as input for Bi-LSTM}
\vspace{-5mm}
\end{table*}

\begin{table*}[]
\label{table8}
\centering
\captionsetup{justification=centering}
\begin{tabular}{|c|c|c|c|c|}
\hline
\textbf{\begin{tabular}[c]{@{}c@{}}Word Embedding \\ Model\end{tabular}}                       & \textbf{\begin{tabular}[c]{@{}c@{}}Knowledge\\ Concepts\end{tabular}} & \textbf{Precision} & \textbf{Recall} & \textbf{\begin{tabular}[c]{@{}c@{}}Macro\\ F1 Score\end{tabular}} \\ \hline
\multirow{4}{*}{\begin{tabular}[c]{@{}c@{}}Fasttext on \\ Post Descriptions\end{tabular}}      & Emoji Names                                                           & 60.51              & 59.76           & 59.74                                                             \\ \cline{2-5} 
                                                                                               & Emoji Senses                                                          & 61.29              & 59.79           & 60.19                                                             \\ \cline{2-5} 
                                                                                               & Emoji Sense Definitions                                               & 61.11              & 60.13           & 60.50                                                             \\ \cline{2-5} 
                                                                                               & Processed Emoji Sense Definitions                                     & 60.48              & 60.03           & 60.02                                                             \\ \hline
\multirow{4}{*}{\begin{tabular}[c]{@{}c@{}}Pretrained Fasttext \\ on Wiki Corpus\end{tabular}} & Emoji Names                                                           & 54.28              & 53.85           & 53.85                                                             \\ \cline{2-5} 
                                                                                               & Emoji Senses                                                          & 53.87              & 53.48           & 53.19                                                             \\ \cline{2-5} 
                                                                                               & Emoji Sense Definitions                                               & 54.20              & 53.15           & 53.02                                                             \\ \cline{2-5} 
                                                                                               & Processed Emoji Sense Definitions                                     & 54.87              & 53.16           & 53.51                                                             \\ \hline
\end{tabular}
\vspace{4mm}
\caption{Accuracies of emotion classification task using Bag of words technique combining text and emoji as input to learn description embedding using fasttext trained word embeddings on the set of captions}
\vspace{-5mm}
\end{table*}

\begin{table*}[]
\label{table9}
\centering
\captionsetup{justification=centering}
\begin{tabular}{|c|c|c|c|c|}
\hline
\textbf{\begin{tabular}[c]{@{}c@{}}Mechanism\\  for Text\end{tabular}} & \textbf{\begin{tabular}[c]{@{}c@{}}Knowledge\\ Concepts\end{tabular}} & \textbf{Precision} & \textbf{Recall} & \textbf{\begin{tabular}[c]{@{}c@{}}Macro\\ F1 Score\end{tabular}} \\ \hline
\multirow{4}{*}{Bag Of Words Approach}                                 & Emoji Names                                                           & 61.38              & 60.45           & 60.43                                                             \\ \cline{2-5} 
                                                                       & Emoji Senses                                                          & 61.94              & 59.98           & 60.60                                                             \\ \cline{2-5} 
                                                                       & Emoji Sense Definitions                                               & 61.00              & 60.11           & 60.19                                                             \\ \cline{2-5} 
                                                                       & Processed Emoji Sense Definitions                                     & 61.30              & 60.63           & 60.88                                                             \\ \hline
\multirow{4}{*}{Attention Mechanism}                                   & Emoji Names                                                           & 64.68              & 61.66           & 62.54                                                             \\ \cline{2-5} 
                                                                       & Emoji Senses                                                          & 64.71              & 61.53           & 62.34                                                             \\ \cline{2-5} 
                                                                       & Emoji Sense Definitions                                               & 64.96              & 61.42           & 62.12                                                             \\ \cline{2-5} 
                                                                       & Processed Emoji Sense Definitions                                     & 63.69              & 61.54           & 61.64                                                          \\ \hline
\end{tabular}
\vspace{4mm}
\caption{Accuracies of emotion classification task considering text (without emoji) and emoji features as separate input layers}
\vspace{-5mm}
\end{table*}

\begin{table*}
\label{table10}
\centering
\captionsetup{justification=centering}
\begin{tabular}{|c|c|c|c|}
\hline
\textbf{\begin{tabular}[c]{@{}c@{}}Knowledge\\ Concepts\end{tabular}} & \textbf{Precision} & \textbf{Recall} & \textbf{\begin{tabular}[c]{@{}c@{}}Macro\\ F1 Score\end{tabular}} \\ \hline
Emoji Names                                                           & 70.23              & 69.25           & 69.42                                                             \\ \hline
Emoji Senses                                                          & 73.79              & 70.25           & 71.98                                                             \\ \hline
Emoji Sense Definitions                                               & 71.56              & 69.98           & 70.49                                                             \\ \hline
Processed Emoji Sense Definitions                                     & 72.23              & 70.26           & 70.78                                                             \\ \hline
\end{tabular}
\vspace{4mm}
\caption{Accuracies of emotion classification task using attention mechanism combining text and emoji as input for Bi-LSTM}
\end{table*}

\subsection{Combining textual, visual and emoji features}

\begin{figure}[H]
\label{fig7}
\centering
\captionsetup{justification=centering}
\includegraphics[width=1.0\linewidth]{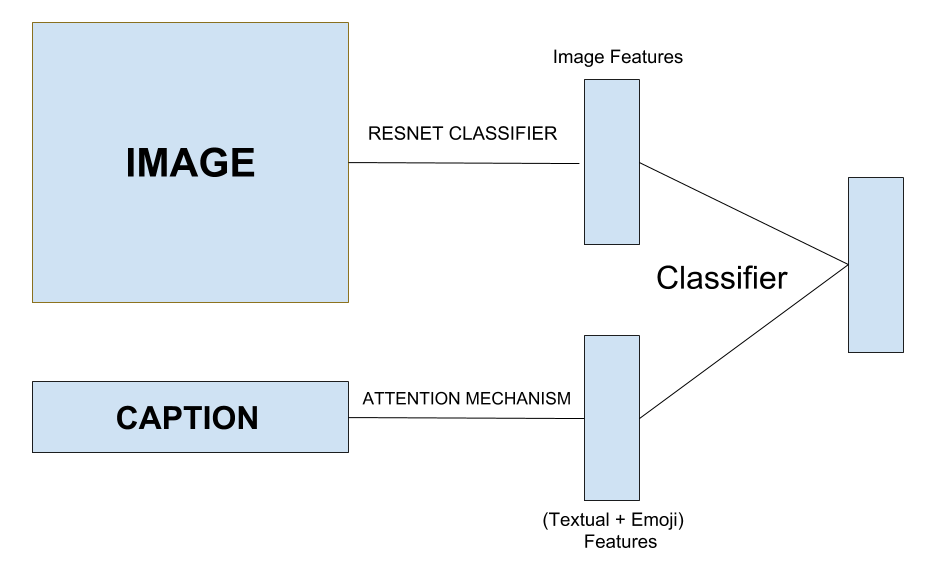}
\caption{Multimodal emotion classification approach using both caption and image}
\end{figure}

Barbieri et al. \cite{barbieri2018multimodal} showed that there would be an increase in prediction accuracies if both visual and textual features are combined. We investigated the same in the context of multimodal emotion classification using SOTA model architectures, i.e., ResNet-152 for learning image features,  attention mechanism to learn features from text and merge these features to form a hidden layer. We then use a softmax layer on top of this for classification. We report the classification accuracies using this model architecture in \hyperref[table10]{Table 10}. \hyperref[fig7]{Figure 7} shows the model architecture.

\section{Discussion}

We presented extensive experiments to check the accuracies of different approaches and importance of different modalities towards emotion classification. Researchers have identified the importance to study emoji for social NLP tasks considering its usage on social media platforms. Hence we considered emojis as an addition feature to learn the emotional intent from a social media post. We have first looked at how important visual features are to study user emotions.  We used ResNet architecture developed by He et al. \cite{He2015} to learn features from images. We had observed a very low F score of about 30\% using ResNet-152 model architecture which is currently the state of the art for image classification. This low accuracy can be explained by the presence of captioned photos which may confuse the model while learning image features for emotion classification \cite{hu2014we}.

We then looked at the importance of textual features for emotion classification. Recent research has identified the importance of textual features over visual features for task involving social media data \cite{barbieri2018multimodal}. The same has been observed in the case of emotion classification. A caption may contain emoji or text, hence we looked at the importance of these modalities for emotions classification individually. Unlike Twitter where users are restricted to a caption of character length 300, instagram allows users to use a caption of length 2200 which can give room to about 350 words. In our dataset, we have observed that the maximum characters in the caption is nearly 1948,  maximum word count of a caption is 423 words, and the average number of words in the caption is about 44 words. We then looked at the variation of accuracies with  input sentence length. We observed that the maximum classification accuracy at a input sentence length of 80 words/caption. Further increasing the sentence length did not affect the classification accuracy.

We then looked at the importance of the third modality-- emoji features, we used the different emoji knowledge concepts extracted from EmojiNet and trained a neural network. We used the vector average of the emojis embeddings used in each post as the feature vector and input of the neural network. Considering the use of \includegraphics[height=1em]{1268.png} in different contexts, the emoji \includegraphics[height=1em]{1268.png} is the most frequently used emoji to express Love or Joy. Hence using only emoji features for emotion classification is not a good idea and the same is observed in the results. The classification accuracy using emoji features is about 32.83\% which is very low compared to textual features. 

We then looked at the importance of emojis in addition to text for the emotion classification task. We used three different model architectures., The first was the bag of words model for the whole caption (emojis embedded in between text). The second being attention mechanism with whole caption as input (emojis embedded in between text), where we consider an embedding layer consisting of both textual and emoji embeddings to train an attention model. In the third architecture, emojis and text are considered as two different features and fed through different input layers. The accuracies have been found to be better if emojis embedded with text is considered as sequential input to the attention model. This can be related to a study by Kevin Cohn which says that emojis on social media are being used as a language\footnote{\url{https://bit.ly/2UbmTRG}} and hence a Bi-LSTM model would give better accuracies for sequential data. For example consider the caption ``My \includegraphics[height=1em]{1268.png} for life!!", here \includegraphics[height=1em]{1268.png} is used in context of ``love". If both text and emoji are considered as different inputs, this would confuse the model since \includegraphics[height=1em]{1268.png} can be used in different contexts (as seen in Table 2), this decreases the classification accuracy. This is the reason resulting in high accuracy when both text + emoji is considered as sequential input to attention model.

Finally, we combined the three modalities -- textual, visual, and emoji. \hyperref[table10]{Table 10} reports the classification accuracies observed using different emoji knowledge concepts wherethe caption (text + emoji) is sent through attention mechanism and image is sent through ResNet-152 model. We then merge the outputs of these layers and train a softmax on top of this for classification. This combination of all the modalities results in better classification accuracy.

\section{Conclusion and Future Work}

In this paper, we explored the usage of different emojis in different emotional contexts and the importance of different modalities towards emotion classification task. We have presented a multimodal emotion classification approach which makes use of all modalities -- emoji, textual and visual features. We have further shown that combining all modalities can outperform state of the art unimodal approaches (based on only on textual or visual or emoji contents). We also observed better accuracy for the emotion classification task when the caption (emoji and text) is considered as sequential input compared to accuracy when used as different input features. As a future work and with Felbo et al. \cite{felbo2017using} as reference, we plan to work on building transfer learning approaches using pre-trained classifiers to learn the emotional features from visual and textual contents towards emotion classification task. We also plan to evaluate our approaches using human annotated test set.

\bibliographystyle{ACM-Reference-Format}


\end{document}